\documentclass{article}

\usepackage[nonatbib,preprint]{nips_2018}

\usepackage[utf8]{inputenc} 
\usepackage[T1]{fontenc}    
\usepackage{hyperref}       
\usepackage{color}
\usepackage{url}            
\usepackage{booktabs}       
\usepackage{amsfonts}       
\usepackage{nicefrac}       
\usepackage{microtype}      
\usepackage{graphicx} 
\usepackage{subfigure}
\usepackage{amsmath}
\usepackage{tablefootnote}

\title{Large-Batch Training for LSTM and Beyond}

\author{
  Yang You$^2$\thanks{Work done as part of internship at Google Brain.}, Jonathan Hseu$^1$, Chris Ying$^1$, James Demmel$^2$, Kurt Keutzer$^2$, Cho-Jui Hsieh$^{1,3}$\\
  Google$^1$, UC Berkeley$^2$, UCLA$^3$\\
}

\begin{document}

\maketitle

\begin{abstract}
Large-batch training approaches have enabled researchers to utilize large-scale distributed processing and greatly accelerate deep-neural net (DNN) training.
For example, by scaling the batch size from 256 to 32K \cite{you2017scaling}, researchers have been able to reduce the training time of ResNet50 on ImageNet from 29 hours to 2.2 minutes \cite{ying2018image}.
However, there are three problems in current large-batch research:
(1) Although recurrent neural net (RNN) approaches like LSTM \cite{hochreiter1997long} have been widely used in many real-world applications, current large-batch research is principally focused on Convolutional Neural Nets (CNN).
(2) Even for CNNs, there is no \textit{automated} technique for  extending the batch size beyond 8K. Instead it requires significant parameter tuning.
(3) To keep the variance in the gradient expectation constant, theory suggests that a Sqrt Scaling scheme \cite{krizhevsky2014one} should be used in large-batch training. Unfortunately, there are not many successful applications using the Sqrt Scaling scheme.
In this paper, we propose a new approach called linear-epoch gradual-warmup (LEGW) for better large-batch training. 
With LEGW, we are able to conduct large-batch training for both CNNs and RNNs with the Sqrt Scaling scheme. 
LEGW enables Sqrt Scaling scheme to be useful in practice and as a result we achieve much better results than the Linear Scaling learning rate scheme (Figure \ref{fig:legw_vs_ls_resnet50}).
For LSTM applications, we are able to scale the batch size by a factor of 64  without losing accuracy and without tuning the hyper-parameters.
For CNN applications, LEGW is able to achieve the same accuracy even as we scale the batch size to 32K. LEGW works better than previous large-batch auto-tuning techniques (Figure \ref{fig:legw_vs_ls_resnet50}).
LEGW achieves a 5.3$\times$ average speedup over the baselines for four LSTM-based applications on the same hardware.
We also provide some theoretical explanations for LEGW.
\end{abstract}

\section{Introduction}
Speeding up Deep Neural Network (DNN) training is important because it can improve the productivity of machine learning researchers and developers.
Since the acceleration of training through exploiting  model parallelism is limited,  current research principally focuses on data parallelism.
Specifically, a large-batch training approach has enabled us to successfully exploit large-scale distributed processing \cite{akiba2017extremely,goyal2017accurate,jia2018highly,li2017scaling,you2017imagenet,ying2018image,osawa2018second}.
For example, by scaling the batch size from 256 to 32K \cite{you2017scaling}, researchers are able to reduce the training time of ResNet50/ImageNet from 29 hours \cite{he2016deep} to 2.2 minutes \cite{ying2018image}.
However, there are three problems with current large-batch approaches:

\begin{itemize}
  \item Although RNN techniques like LSTM \cite{hochreiter1997long} have been widely used, the current large-batch study is mostly focused on CNN applications. On the other hand, adaptive solvers like Adam do not beat well-tuned Momentum SGD for ImageNet training.
  We want to evaluate Adam for large-batch LSTM training.
  \item Even for CNN applications, significant hyper-parameter tuning is required to increase the batch size beyond 8K with no loss in accuracy. For batch sizes lower than 8K, linear scaling usually works well for most applications. However, for batch sizes beyond 8K, even solvers like LARS \cite{you2017scaling} require users to manually tune the hype-parameter (including learning rate, warmup, weight decay, and momentum). 
  \item Prior successful large-batch training relies on a linear scaling scheme \cite{goyal2017accurate}. However, to keep the variance in the gradient expectation constant, theory \cite{krizhevsky2014one} suggests a Sqrt Scaling scheme should be used. Currently there are not many successful large-batch applications using Sqrt Scaling. 
  It will be ideal if we can bridge the gap between theory and practice. 
\end{itemize}

\begin{figure}
\centering
\includegraphics[width=3.9in]{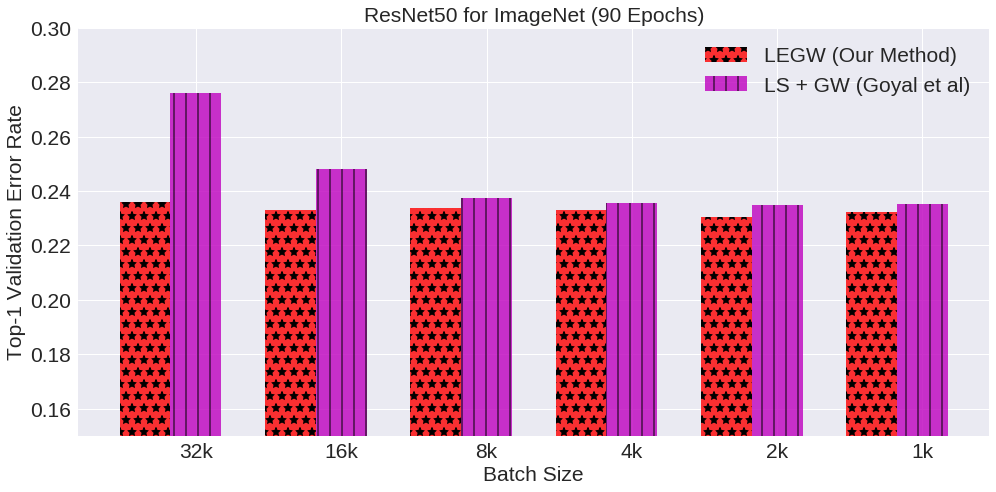}
\caption{\footnotesize \label{fig:legw_vs_ls_resnet50}
LEGW achieves the constant accuracy when we scale up the batch size without tuning the parameters (learning rate, weight decay, and momentum). LEGW works better than previous large-batch tuning techniques \cite{goyal2017accurate}.}
\vspace{-10pt}
\end{figure}

To solve these problems, we propose linear-epoch gradual-warmup approach in this paper.
We call this approach {\bf Leg-Warmup} (LEGW).
LEGW enables a Sqrt Scaling scheme in practice and as a result we achieve much better performance than the previous Linear Scaling learning rate scheme.
For the GNMT application (Seq2Seq) with LSTM, we are able to scale the batch size by a factor of 16 without losing accuracy and without tuning the hyper-parameters mentioned above.
For the PTB dataset with LSTM, we are able to scale the batch size by a factor of 32 without losing accuracy and without tuning the hyper-parameters.
Beyond RNN applications, we also successfully applied LEGW in ImageNet training with ResNet50.
When used together with the LARS solver, LEGW is able to achieve the constant accuracy when we scale the batch size to 32K. LEGW works better than previous large-batch tuning techniques (Figure \ref{fig:legw_vs_ls_resnet50}).
We also give some theoretical insights to explain why LEGW works well in large-batch training.
LEGW achieves a 5.3$\times$ average speedup over the baselines for 4 LSTM-based applications on the same hardware.

\section{Background and Related Work}
\label{sec:background}

\subsection{Data-Parallelism Mini-Batch SGD}
Let us refer to $w$ as the DNN weights, $X$ as the training data, $n$ as the number of samples in $X$, and $Y$ as the labels of $X$.
Let us also denote $x_i$ as a sample of $X$ and $l(x_i, w)$ as the loss computed by $x_i$ and its lable $y_i$ ($i \in \{1, 2, ..., n$\}). 
A typical loss function is cross-entropy.
The goal of DNN training is to minimize the loss defined in Equation (\ref{eq:loss}).

\begin{equation}
    L(w) = \frac{1}{n} {\sum}_{i=1}^{n} l(x_i, y_i, w)
    \label{eq:loss}
\end{equation}

At the $t$-th iteration, we use forward and backward propagation to get the gradients of weights based on the loss. Then we use the gradients to update the weights, which is shown in Equation (\ref{eq:update1}):

\begin{equation}
    w_{t+1} = w_t - \eta \nabla l(x_i, y_i, w)
    \label{eq:update1}
\end{equation}

where $\eta$ is the learning rate. This method is called as Stochastic Gradient Descent (SGD). Usually, people do not use a single sample to compute the loss and the gradients. They use a batch of samples at each iteration. Let us refer to the batch of sample at $t$-th iteration as $B_t$. The size of $B_t$ is $b$. Then we update the weights based on Equation (\ref{eq:update2}).

\begin{equation}
    w_{t+1} = w_t - \frac{\eta}{b} {\sum}_{x \in B_t}\nabla l(x, y, w)
    \label{eq:update2}
\end{equation}

This method is called as Mini-Batch SGD. To simplify the notation, we define the gradient estimator as $\nabla w_t :=  \frac{1}{b} {\sum}_{x \in B_t}\nabla l(x, y, w)$ and the updating rule in Equation (\ref{eq:update3}) can be denoted as
\begin{equation}
    w_{t+1} = w_t - \eta \nabla w_t.
    \label{eq:update3}
\end{equation}
%

\subsection{Large-Batch Training Difficulty}
Increasing the batch size allows us to scale to more machines without reducing the workload on each machine \cite{shallue2018measuring}. 
On modern architecture like TPUs, reducing the workload often leads to a lower efficiency. 
However, when we increase the batch size after a certain point (e.g. 1024) without carefully tuning the hyper-parameters, the algorithm usually suffers from slow convergence. 
The test accuracy of the converged solution becomes significantly lower than the baseline \cite{goyal2017accurate,hoffer2017train,keskar2016large,li2014efficient}.
Keskar et al \cite{keskar2016large} suggested that there is a generalization problem for large-batch training. 
Hoffer et al \cite{hoffer2017train} and Li et al \cite{li2014efficient} suggests that training longer will help algorithm to generalize better and keep the accuracy. On the other hand, Goyal et al \cite{goyal2017accurate} can scale the batch size to 8K without losing accuracy.

\subsection{Large Batch Training Techniques}
When we increase the batch size ($B$), we need to increase the initial LR to prevent losing accuracy \cite{goyal2017accurate}. There are two rules of increasing the initial LR:

{\bf Sqrt Scaling Rule \cite{krizhevsky2014one}}. When we increase the batch size by $k$ times, we should increase the LR by $\sqrt{k}$ times to keep the variance of the gradient estimator constant.

{\bf Linear Scaling Rule \cite{krizhevsky2014one}}: When we increase the batch size by $k$ times, we should increase the LR by $k$ times based on the assumption that $\nabla l (x, y, w_t) \approx \nabla l (x, y, w_{t+j})$, where $j < B$. 

{\bf Warmup Scheme \cite{goyal2017accurate}} Usually, under linear scaling rule, $k \eta$ is exetremely large, which may make the algorithm diverge at the beginning. Therefore, people set the initial LR to a small value and increase it gradually to $k \eta$ in a few epochs (e.g. 5 or 10). This method is called {\bf Gradual Warmup Scheme}. 

Krizhevsky \cite{krizhevsky2014one} reported 1 percent loss in accuracy when he increased the the batch size from 128 to 1024. 
Iandola et al \cite{iandola2016firecaffe} also scaled the batch size to 1K for AlexNet and GoogLeNet.
Li \cite{li2017scaling} used a batch of 5120 for ResNet-101 to train Imagenet dataset on 160 GPUs. 
Goyal et al \cite{goyal2017accurate} scaled the batch size to 8K for ImageNet training with ResNet-50. 
The LARS algorithm \cite{you2017scaling} was proposed to scale the batch size to 32K for ImageNet training. The LARS algorithm was implemented on 2048 Intel KNL chips and finished the ImageNet/ResNet-50 training in 14 minutes \cite{you2017imagenet}.
The LARS algorithm was also implemented on TPU-v3 Pod to finish the ImageNet/ResNet-50 training in 2.2 minutes \cite{ying2018image}.
Codreanu et al \cite{codreanu2017scale} scaled DNN training on 1024 SkyLake CPUs and finished ImageNet training with ResNet50 in 44 minutes.
Akiba et al \cite{akiba2017extremely} scaled the batch size to 32K and finish the ImageNet training with ResNet50 in 15 minutes. However, their baseline's accuracy was missing.
Jia et al \cite{jia2018highly} combined LARS algorithm with mixed-precision training \cite{micikevicius2017mixed} and finished the ImageNet training with ResNet50 in 8.6 minutes.
The other related directions include K-FAC \cite{martens2015optimizing} and dynamic batch size \cite{devarakonda2017adabatch,smith2017don}.

\section{Linear-Epoch Gradual Warmup (LEGW)}
The warmup technique has been successfully applied in the CNN applications \cite{goyal2017accurate,you2017scaling}.
However, most of the RNN implementations did not use warmup techniques.
On the other hand, warmup has become an additional parameter that requires developers to tune, which further increases the efforts of DNN system implementation.
We propose the Linear-Epoch Gradual Warmup (LEGW or Leg-Warmup) scheme.
When we increase the batch size by $k$ times, we also increase the warmup epochs by $k$ times.
The intuition is that larger batch size usually needs a large learning rate (LR).
However, larger LR may make the training algorithm easier to diverge because the gradient changes dramatically in the beginning of neural network training.
We use longer warmup to avoid the divergence of larger LR.

\begin{figure}[ht]
\centering
\renewcommand{\thesubfigure}{\thefigure.\arabic{subfigure}}
\subfigure[]{\includegraphics[width=3.9in]{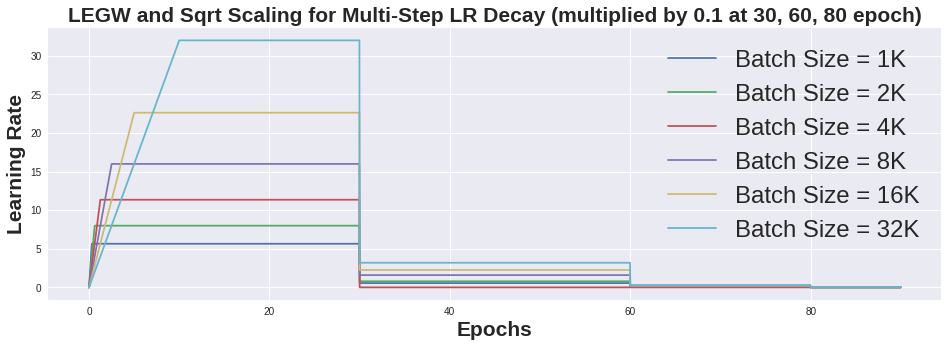}
\label{fig:legw_visualize1}}
\subfigure[]{\includegraphics[width=3.9in]{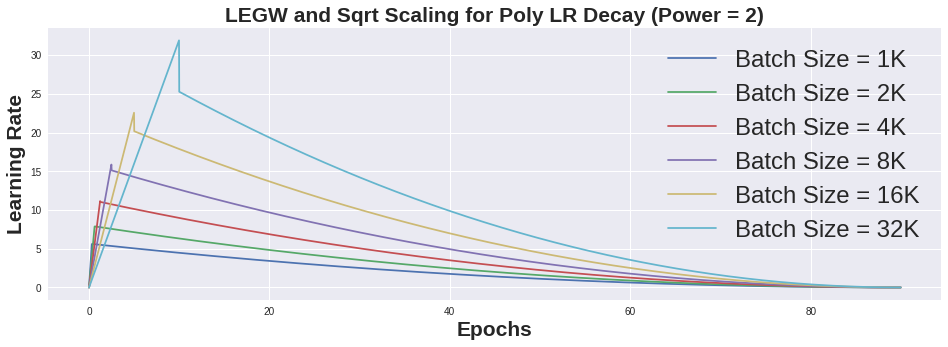}
\label{fig:legw_visualize2}}
\vspace{-5pt}\caption{\footnotesize Example of LEGW for ImageNet training with ResNet50. The figures show the example of using multi-step learning rate decay and polynomial learning rate decay (power=2.0).\vspace{-10pt}}
\label{fig:legw_visualize}
\end{figure}

\subsection{Sqrt Learning Rate Scaling}
To keep the variance in the gradient expectation constant, theory \cite{krizhevsky2014one} suggests that Sqrt Scaling scheme should be used in large-batch training.
In practice, however, researchers observe that Linear Scaling scheme works better than Sqrt Scaling  scheme \cite{goyal2017accurate,krizhevsky2014one,li2017scaling,you2017imagenet}.
The constant-epoch warmup scheme was used together with Linear Scaling in previous applications.
For example, Goyal et al. \cite{goyal2017accurate} manually set the warmup length at five epochs.
The efficiency of Linear Scaling only works up to 8K batch size, although researchers are able to scale the batch size to 32K with signifiant hyper-parameter tuning (tuning learning rate, warmup, weight decay and momentum for different batch sizes).
With LEGW, the Sqrt Scaling scheme can work well in practice, and is able to match the expectations the theoretical analysis.
The results are shown in Section \ref{sec:results}.

\begin{figure}[ht]
\centering
\vspace{-10pt}\renewcommand{\thesubfigure}{\thefigure.\arabic{subfigure}}
\subfigure[SGD with batch size 512. ]{\includegraphics[width=3.6in]{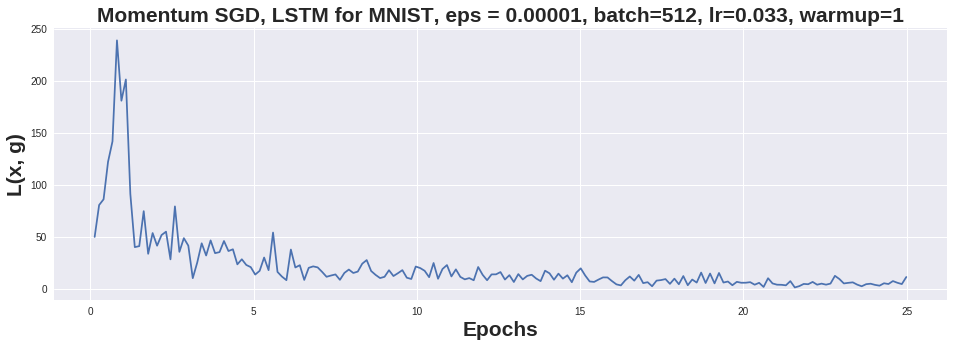}
\label{fig:legw_explain1}}
\subfigure[SGD with batch size 1K.]{\includegraphics[width=3.6in]{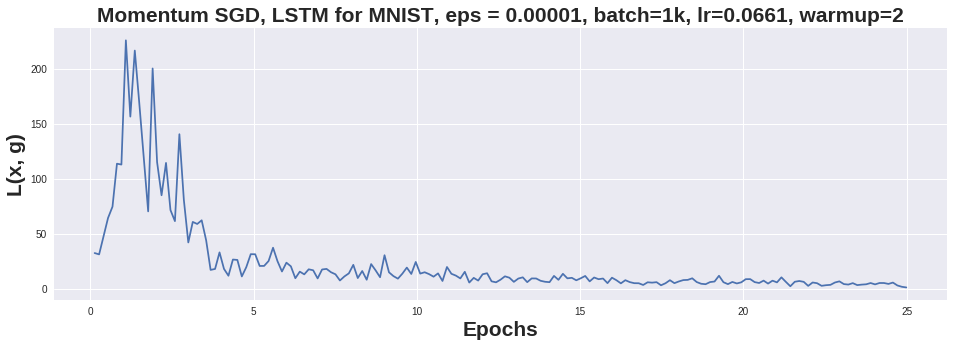}
\label{fig:legw_explain2}}
\subfigure[SGD with batch size 2K.]{\includegraphics[width=3.6in]{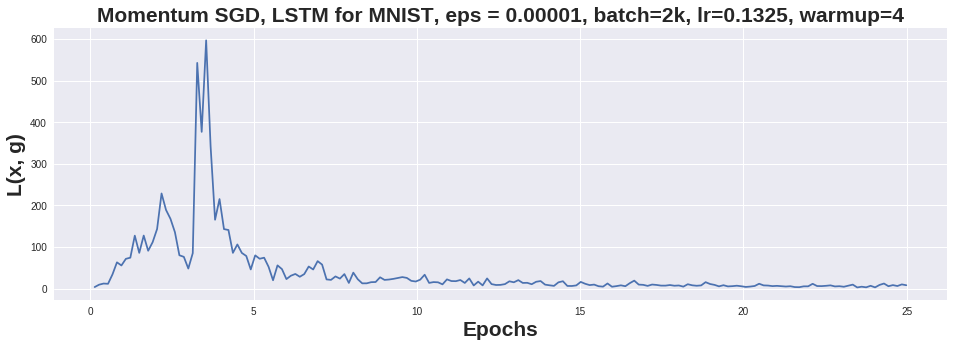}
\label{fig:legw_explain3}}
\subfigure[SGD with batch size 4K.]{\includegraphics[width=3.6in]{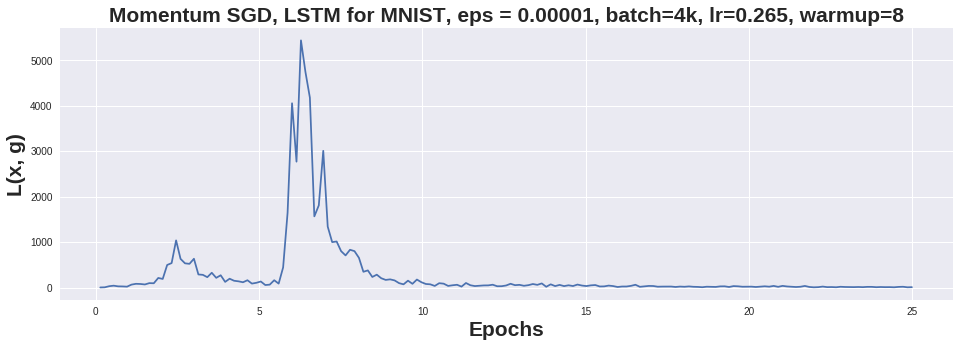}
\label{fig:legw_explain4}}
\vspace{-10pt}\caption{\footnotesize The approximation of Lipchitz constant for different batch sizes.}
\label{fig:legw_explain}
\end{figure}

\subsection{Illustration of LEGW}
To illustrate how LEGW works with commonly-used LR decay schemes, we use ImageNet training with ResNet50.
First, we use multi-step LR decay (essentially the same with exponential decay) scheme (Figure \ref{fig:legw_visualize}.1).
The baseline uses a batch size of 1K and an initial learning rate of 2$^{2.5}$.
In the initial 0.3125 epochs (35,190 iterations), LEGW gradually increases LR from 0 to 2$^{2.5}$.
From 0.3125 epoch to 30th epoch, LEGW uses the constant learning rate of 2$^{2.5}$.
From 30th epoch to 60th epoch, LEGW uses the constant learning rate of 0.1$\times$2$^{2.5}$.
From 60th epoch to 80th epoch, LEGW uses the constant learning rate of 0.01$\times$2$^{2.5}$.
From 80th epoch to 90th epoch, LEGW uses the constant learning rate of 0.001$\times$2$^{2.5}$.
When we scale the batch size from 1K to 2K, LEGW increases the learning rate from 2$^{2.5}$ to 2$^{3.0}$ based on the Sqrt Scaling scheme.
For batch size of 2K, LEGW warms up the learning rate gradually in the initial 0.625 epochs. 
In the same way, LEGW reduces the learning rate by multiplying it by 0.1 at 30th, 60th, and 80th epoch.
The same idea applies when we further scale up the batch size.
The details are shown in Figure \ref{fig:legw_visualize}.1.
Some users may feel there are too many parameters in multi-step LR decay scheme.
The users may need to decide in which epoch to decay the learning rate, and how much the learning rate should be reduced each time.
Another commonly-used scheme is polynomial decay or poly decay.
Let us use $p$ to denote the power of poly decay, $\eta$ the initial LR, $i$ the current iteration, and $I$ the total number of iterations.
The learning rate of iteration $i$ is $\eta \times (1 - i/I)^p$.
The baseline uses a batch size of 1K and an initial learning rate of 2$^{2.5}$.
In the initial 0.3125 epoch (35,190 iterations), LEGW gradually increases LR from 0 to 2$^{2.5}$.
From the 0.3125 epoch to the 90th epoch, LEGW uses a LR of $2^{2.5} \times (1 - i/I)^2$.
The same idea applies when we further scale up the batch size.
The details are shown in Figure \ref{fig:legw_visualize}.2.

\subsection{Minimal Tuning Effort}
By using LEGW, the users do not need to manually tune the LR for different batch sizes.
For example, the users only need to tune the hyper parameters of a baseline (e.g. batch size = 256).
Then, if the user scales up the batch size by $k$ times, they only need to increase the learning rate by $\sqrt{k}$ times and the number of warmup epochs $k$ times.
On the other hand, the users may also choose and tune the hyper-parameters of a large-batch case (e.g 32K) and then use LEGW to automatically get the LR schedule for smaller-batch cases.
Running a large-batch case is much faster than running a baseline (if the users have enough computational resources).
So tuning large batch maybe faster than tuning the small batch.
In this way, when a user scales down the batch size by $k$ times, they only need to decrease the learning rate by $\sqrt{k}$ times and the number of warmup epochs $k$ times.
\vspace{-5pt}
\section{Explanation of LEGW}
\vspace{-5pt}In general, it is hard to prove that a specific learning rate schedule works. However, some experimental findings  on the change of local Lipschitz constant during iterations partially explain why LEGW works better than previous methods. 

Consider the update along the gradient direction $g$. Assume the update is $x \leftarrow x - \eta g$, the question is: how to choose step size $\eta$? One classical idea is to form a second order approximation around current solution $x$. $f(x+\Delta) \approx$ 
\begin{equation}
     \tilde{f}(x+\Delta) := \{f(x) + \Delta^T \nabla f(x) + \frac{1}{2}\Delta^T \nabla^2 f(x) \Delta \}, 
\end{equation}
and then find $\Delta$ to minimize the approximation function. If we assume $\Delta$ is in the form of $-\eta g$, then the optimal $\eta^*$ is
\begin{align*}
    \arg\min_{\eta} \tilde{f}(x- \eta g) 
    = \frac{1}{\|g^T \nabla^2 f(x) g\|/\|g\|^2}:=\frac{1}{L(x, g)}.  
\end{align*}
Therefore, ideally the step size should be inversely proportional to $L(x, g)$. Moreover, it is known that the update $-\eta g$ will decrease the objective function if $\eta < \min_{x'\in S} \frac{1}{L(x, g)} $ within the region $S$. This is also called the local Lipchitz constant along the gradient direction, and $L(x, g)$ can be viewed as its approximation. 
In Figure~\ref{fig:legw_explain}, we plot the values of 
$L(x, g)$ for all the iterations. It is hard to compute $L(x, g)$ exactly since $\nabla^2 f(x)$ involves all the training samples), so we approximate it using a small batch and compute the Hessian-vector product by finite difference.  For the same reason it is hard to apply a second order method exactly, but the plots in the figures show an interesting phenomenon that explains why linear warmup works. We observe that the value of $L(x,g)$ usually has a peak in the early iterations, implying a smaller step size should be used in the beginning (which implies warmup is needed). Furthermore, the peak tends to shift toward right (almost linearly) as  batch size grows. This intuitively explains our linear warm-up strategy: when batch size increases, the warm up should be longer to cover the ``peak region''. 


\section{Experimental Results\label{sec:results}}
In all the comparisons of this paper, different methods will use the same hardware and run the same number of epochs.
The models and datasets are shown in Table \ref{tab:datasets}.

\begin{table}
 \scriptsize
  \caption{\footnotesize The applications we used to evaluate our method.
  }
  \label{tab:datasets}
\centering
    \vspace{3pt}
  \begin{tabular}{|c|c|c|c|c|c|c|}
  \hline
  Model & Dataset & Samples & Metric \& Reference\\
    \hline
    \hline
    1-layer LSTM & MNIST & 60K/10K & 98.7\% accuracy\tablefootnote{https://medium.com/machine-learning-algorithms} \\
   \hline
   PTB-small & PTB & 930K/82K & 116 perplexity\tablefootnote{https://github.com/tensorflow/models/blob/master/tutorials/rnn/ptb/ptb\_word\_lm.py} \\
   \hline
   PTB-large & PTB & 930K/82K & 78 perplexity\tablefootnote{https://github.com/tensorflow/models/blob/master/tutorials/rnn/ptb/ptb\_word\_lm.py} \\
   \hline
   GNMT & wmt16 & 3.5M/3K & 21.8 BLEU\tablefootnote{https://github.com/mlperf/training/tree/master/rnn\_translator} \\
   \hline
   ResNet50 & ImageNet & 1.3M/5K & 75.3\% accuracy\tablefootnote{https://github.com/KaimingHe/deep-residual-networks} \\
   \hline
  \end{tabular}
\end{table}

\subsection{The LSTM applications}
\subsubsection{\bf Handwritten Digits Recognition for MNIST}
We use this MNIST dataset \cite{lecun1998gradient} to train a pure-LSTM model.
We partition each image as 28-step input vectors. The dimension of each input vector is 28-by-1.
Then we have a 128-by-28 transform layer before the LSTM layer, which means the actual LSTM input vector is 128-by-1.
The hidden dimension of LSTM layer is 128.
Thus the cell kernel of LSTM layer is a 256-by-512 matrix.
State-of-the-art single-layer LSTM achieved 97.27\% accuracy for MNIST dataset \cite{neil2016phased,wang2016recurrent,yu2015multi}.
After careful hyper-parameter tuning, we achieved 98.7\% accuracy in 25 epochs training.
We use a momentum solver (momentum=0.9) and constant learning rate.
The baseline's batch size is 128.
Our goal is to scale the batch size to 8K without losing accuracy.
A batch size over 8K on a V100 GPU will get no additional speedup, so we stop at 8K.
The effect of LEGW is shown in Figures \ref{fig:legw_vs_adam},  \ref{fig:legw_vs_tuning} and \ref{fig:legw_vs_tuning_long}.
These figures show that LEGW is able to beat comprehensive tuning and an Adam solver. 

\subsubsection{\bf Language Modeling for PTB Dataset}
The Penn Treebank (PTB) \cite{marcus1993building} dataset selected 2,499 stories from a three year Wall Street Journal (WSJ) collection of 98,732 stories for syntactic annotation. 
The vocabulary has 10,000 words.
After word embedding, the input vector length is 200 and 1500 for PTB-small model and PTB-large model\footnote{https://github.com/tensorflow/models/blob/master/tutorials/rnn/ptb/ptb\_word\_lm.py}, respectively.
The sequence length is 20 and 35 for PTB-small and PTB-large.
Our LSTM model has two layers. The hidden dimensions of both these two layers are 200 for PTB-small and 1500 for PTB-large.
For both layers, the LSTM Cell Kernel is an 400-by-800 matrix for PTB-small and 3000-by-6000 matrix for PTB-large.
We use perplexity to evaluate the correctness of our LSTM model (lower is better).
After a 13-epoch training, PTB-small can achieve a perplexity of 116\footnote{https://github.com/tensorflow/models/blob/master/tutorials/rnn/ptb/ptb\_word\_lm.py}. 
After a 55-epoch training, PTB-large can achieve a perplexity of 78\footnote{https://github.com/tensorflow/models/blob/master/tutorials/rnn/ptb/ptb\_word\_lm.py}.
For PTB-small, we use momentum solver (momentum=0.9) and exponential learning rate decay.
The model uses constant learning rate in the first seven epochs.
Then the learning rate will be decayed by 0.4 after each epoch.
For PTB-large, we use LARS solver \cite{you2017scaling} and poly decay (power=2.0).
The baseline's batch size is 20.
Our goal is to scale the batch size to 640 without increasing perplexity.
The effect of LEGW is shown in Figure \ref{fig:legw_vs_adam}, which is able to beat the Adam solver.
The batch size over 640 will lead to the out-of-memory issue on a V100 GPU, so we stop at 640.

\begin{figure}[ht]
\centering
\vspace{-10pt}\renewcommand{\thesubfigure}{\thefigure.\arabic{subfigure}}
\subfigure[]{\includegraphics[width=2.4in]{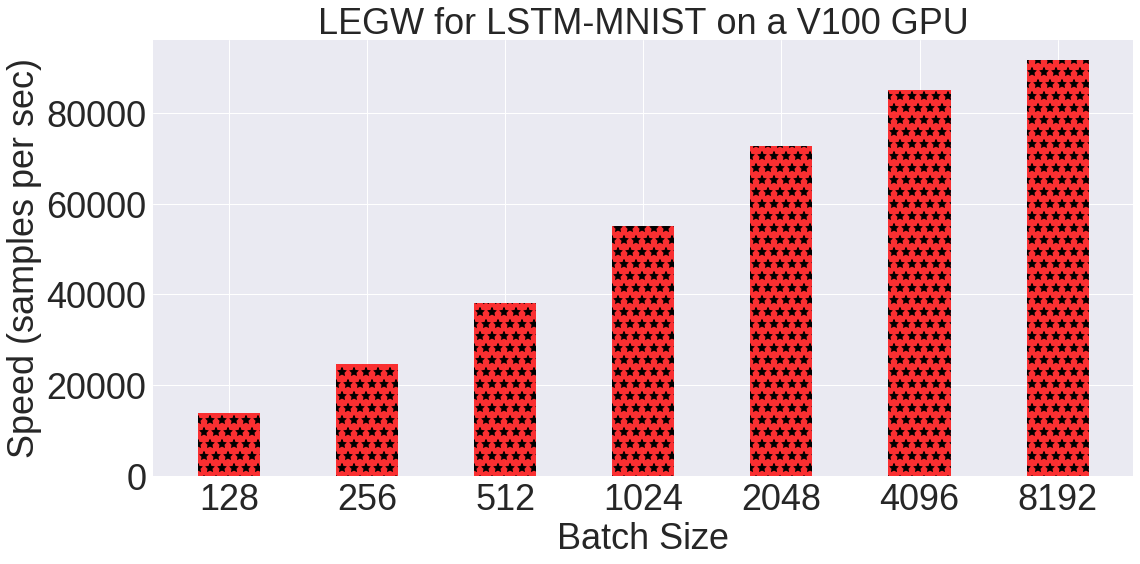}
\label{fig:mnist_speed}}
\subfigure[]{\includegraphics[width=2.4in]{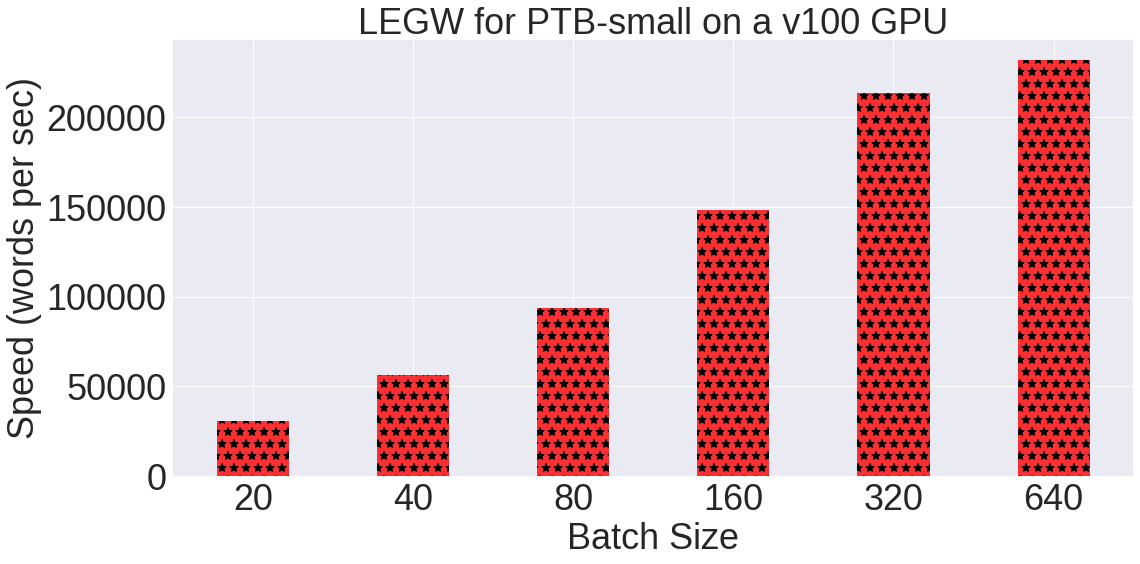}
\label{fig:ptb_small_speed}}
\subfigure[]{\includegraphics[width=2.4in]{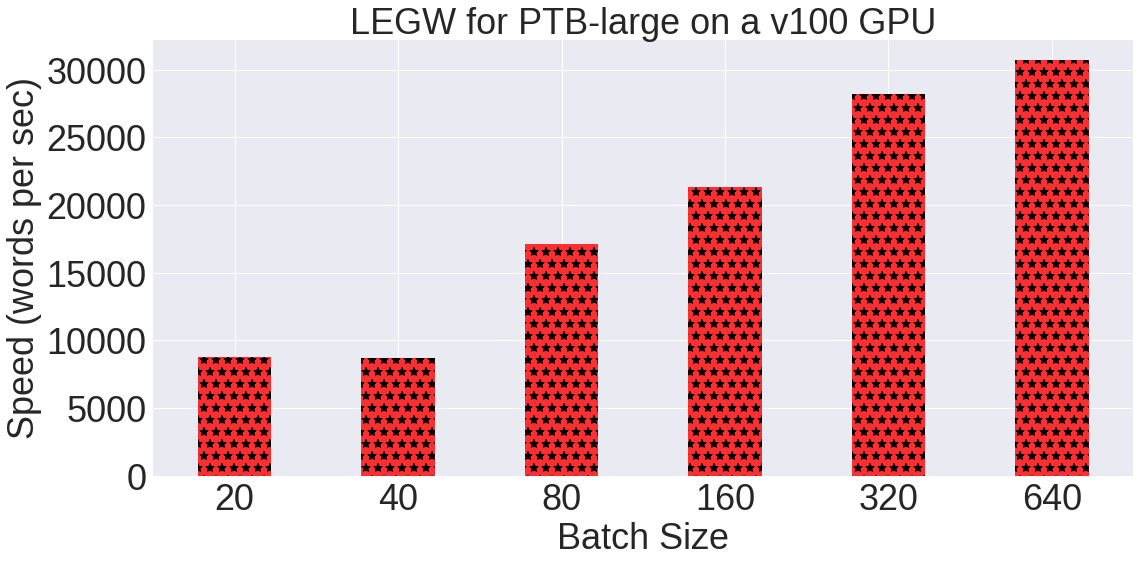}
\label{fig:ptb_large_speed}}
\subfigure[]{\includegraphics[width=2.4in]{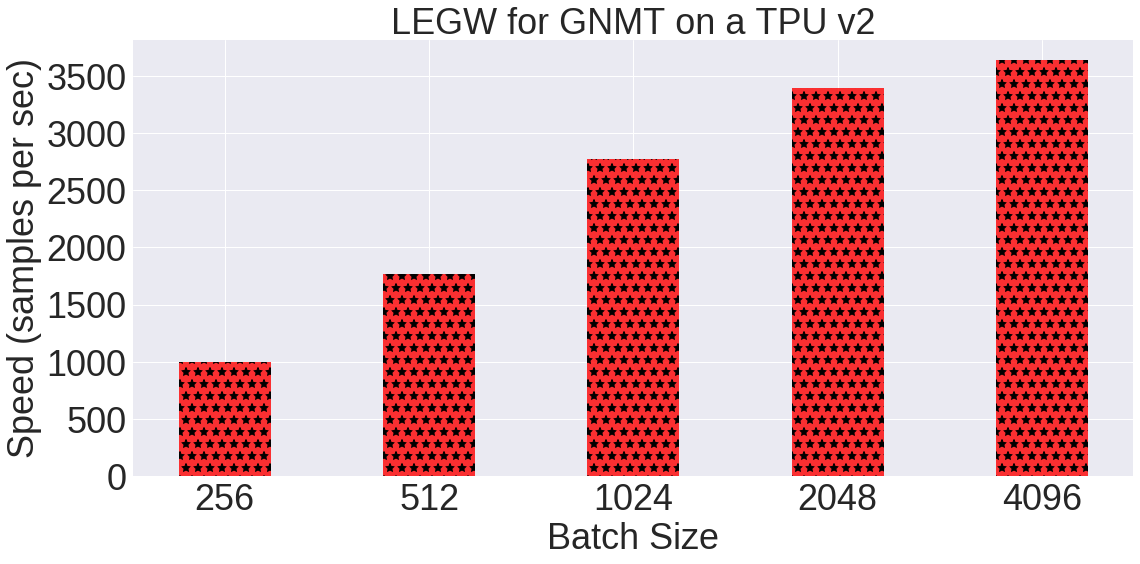}
\label{fig:gnmt_speed}}
\vspace{-10pt}\caption{\footnotesize LEGW approach can scale up the batch size and achieve a higher speed. The left bar is the baseline.
\vspace{-20pt}}
\label{fig:legw_speed}
\end{figure}

\subsubsection{\bf Google Neural Machine Translation (GNMT)}
GNMT or seq2seq \cite{wu2016google,luong17} is a state-of-the-art machine translation technique.
We use WMT16 English-German translation dataset for training.
The encoder and decoder are using shared embeddings.
The encoder includes 4 LSTM layers. The hidden dimension is 1024.
The first layer is bidirectional, the rest are undirectional.
The residual connections start from 3rd layer.
The decoder includes 4 unidirectional LSTM layers with hidden size 1024 and a fully-connected classifier.
The residual connections start from 3rd layer.
We use normalized Bahdanau attention (gnmt\_v2 attention mechanism).
We use BLEU score on newstest2014 dataset as the quality metric (higher is better). The BLEU score is reported by sacrebleu package.
The baseline achieves a BLEU score of 21.80\footnote{https://github.com/mlperf/training/tree/master/rnn\_translator}.
The effect of LEGW is shown in Table \ref{tab:gnmt} and Figure \ref{fig:legw_vs_adam}, which demonstrates that we can scale the batch size to 4K without losing accuracy.
A batch size over 4K will lead to the out-of-memory issue on a TPU, so we stop at 4K.

\subsection{Compared to Adaptive Solvers}
\begin{figure}[ht]
\centering
\renewcommand{\thesubfigure}{\thefigure.\arabic{subfigure}}
\subfigure[]{\includegraphics[width=3.0in]{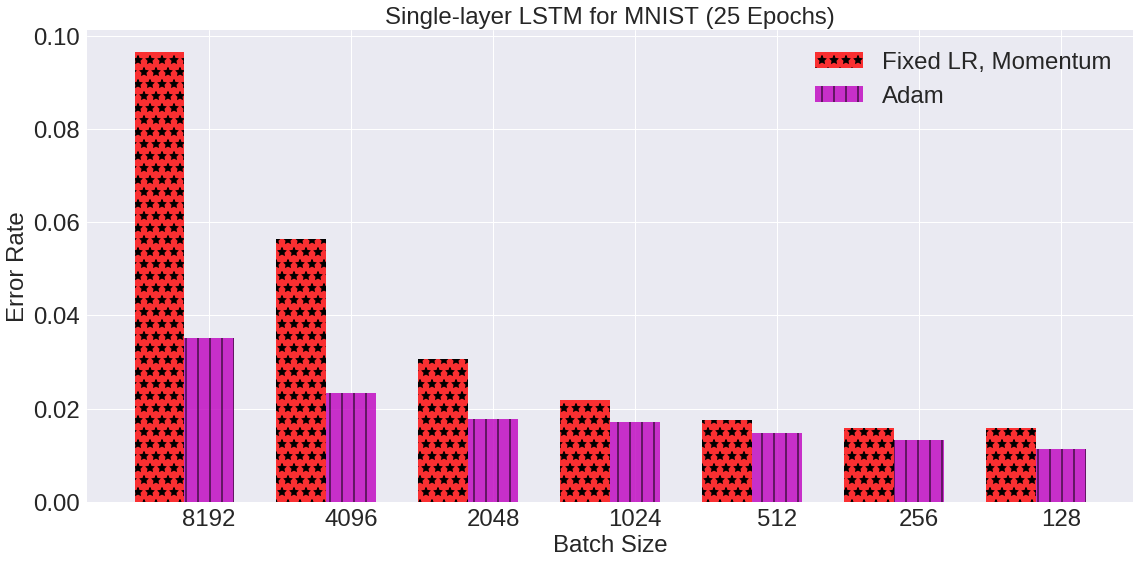}
\label{fig:tuning_vs_adam1}}
\subfigure[]{\includegraphics[width=3.0in]{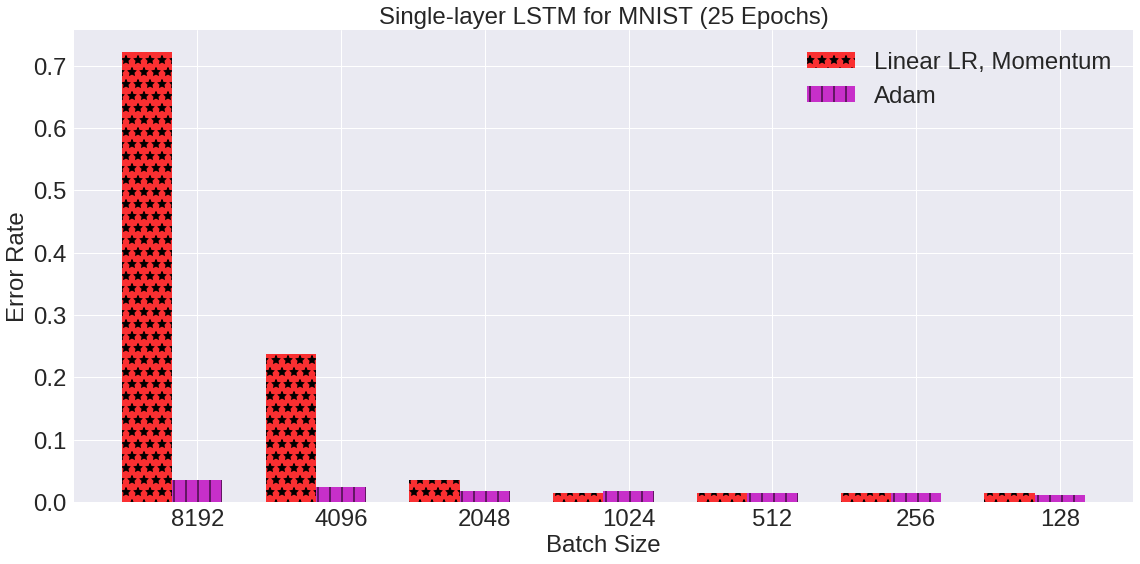}
\label{fig:tuning_vs_adam2}}
\subfigure[]{\includegraphics[width=3.0in]{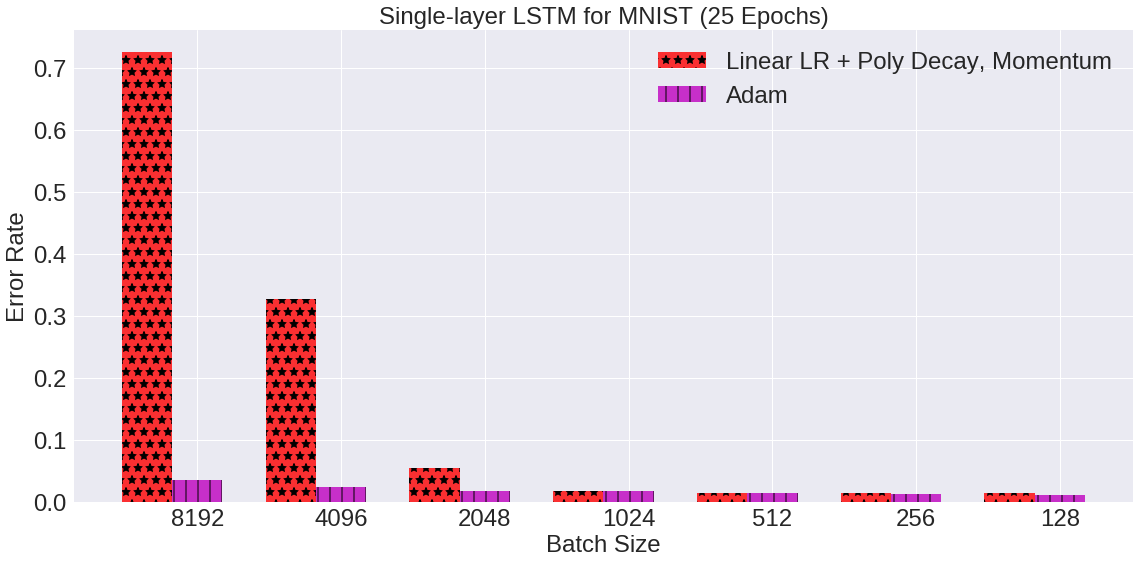}
\label{fig:tuning_vs_adam3}}
\subfigure[]{\includegraphics[width=3.0in]{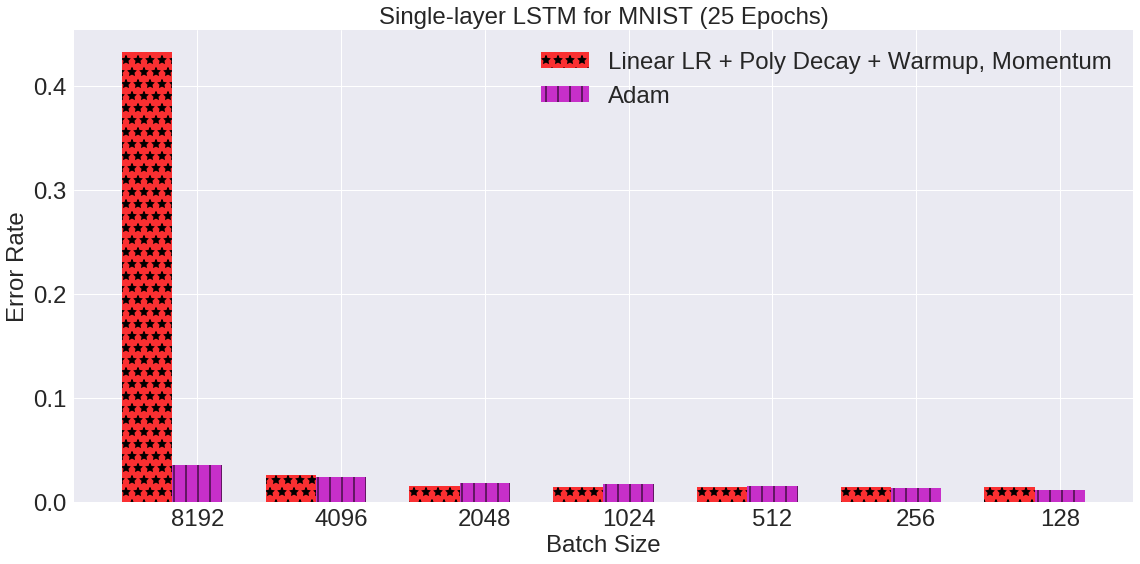}
\label{fig:tuning_vs_adam4}}
\caption{\footnotesize Adam can beat existing tuning techniques. We tune the learning rate for batch size = 128 and refer to it as ${\eta}_0$. Let us also refer to batch size as $B$. In Figure \ref{fig:tuning_vs_adam}.1, all the tuning versions use ${\eta}_0$. In Figure \ref{fig:tuning_vs_adam}.2, all the tuning versions use the linear scaling scheme (i.e. ${\eta}_0 \times B/128$). In Figure \ref{fig:tuning_vs_adam}.3, all the tuning versions use the linear scaling scheme (i.e. ${\eta}_0 \times B/128$) and poly decay with power = 2. In Figure \ref{fig:tuning_vs_adam}.4, all the tuning versions use the linear scaling scheme (i.e. ${\eta}_0 \times B/128$), poly decay with power = 2, and 5-epoch warmup.}
\label{fig:tuning_vs_adam}
\end{figure}

Our goal is to minimize the tuning effort for large-batch training.
To evaluate this we need to pick an adaptive solver as a baseline for comparison.
We fully evaluate a total of seven solvers: SGD \cite{robbins1985stochastic}, Momentum \cite{qian1999momentum}, Nesterov \cite{sutskever2013importance}, Adagrad \cite{duchi2011adaptive}, RMSprop \cite{goodfellow2016deep}, Adam \cite{kingma2014adam}, Adadelta \cite{zeiler2012adadelta}.
We pick Adam and Adadelta as the baseline for adaptive solvers because they do not require the users to input hyper-parameters.
For MNIST and PTB datasets, we observe Adam performs much better than Adadelta.
Moreover, Adam is able to beat the existing tuning techniques (Figure \ref{fig:tuning_vs_adam}). 
We tune the learning rate for batch size = 128 and refer to it as ${\eta}_0$. Let us also refer to batch size as $B$. In Figure \ref{fig:tuning_vs_adam}.1, all the tuning versions use ${\eta}_0$. In Figure \ref{fig:tuning_vs_adam}.2, all the tuning versions use the linear scaling scheme (i.e. ${\eta}_0 \times B/128$). In Figure \ref{fig:tuning_vs_adam}.3, all the tuning versions use the linear scaling scheme (i.e. ${\eta}_0 \times B/128$) and poly decay with power = 2. In Figure \ref{fig:tuning_vs_adam}.4, all the tuning versions use the linear scaling scheme (i.e. ${\eta}_0 \times B/128$), poly decay with power = 2, and 5-epoch warmup.
Thus, we use Adam as the adaptive solver baseline for comparison.

\begin{figure}[ht]
\vspace{-10pt}\centering
\renewcommand{\thesubfigure}{\thefigure.\arabic{subfigure}}
\subfigure[]{\includegraphics[width=3.4in]{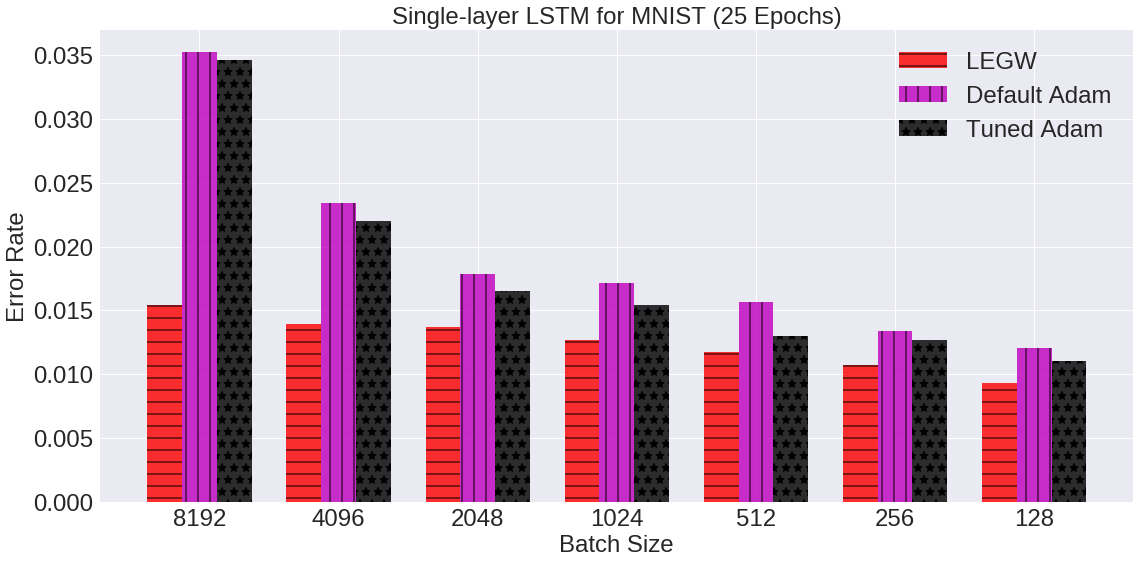}
\label{fig:legw_vs_adam1}}
\subfigure[]{\includegraphics[width=3.4in]{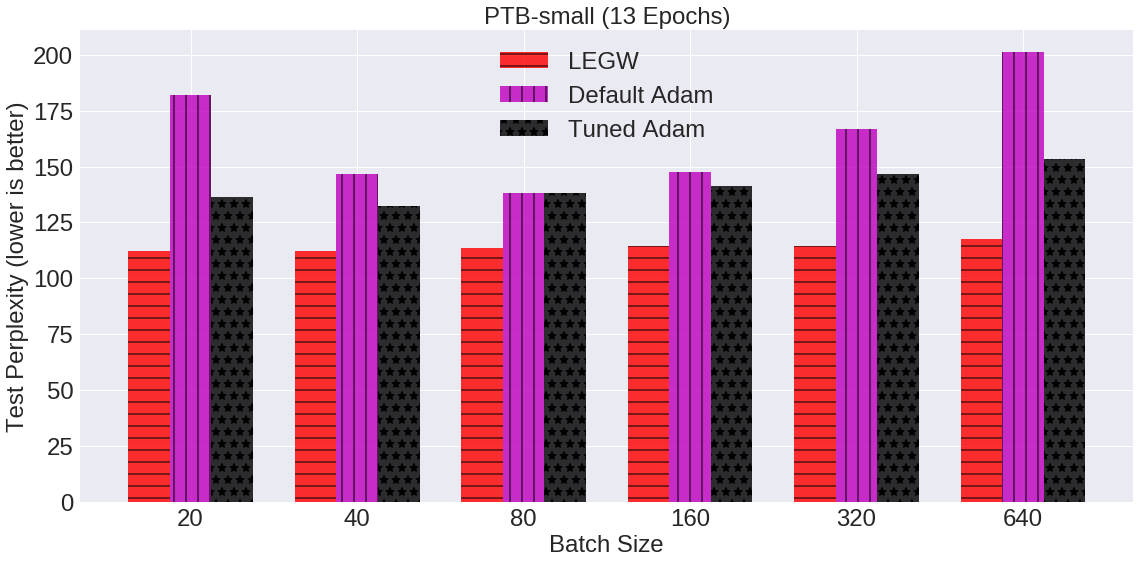}
\label{fig:legw_vs_adam2}}
\subfigure[]{\includegraphics[width=3.4in]{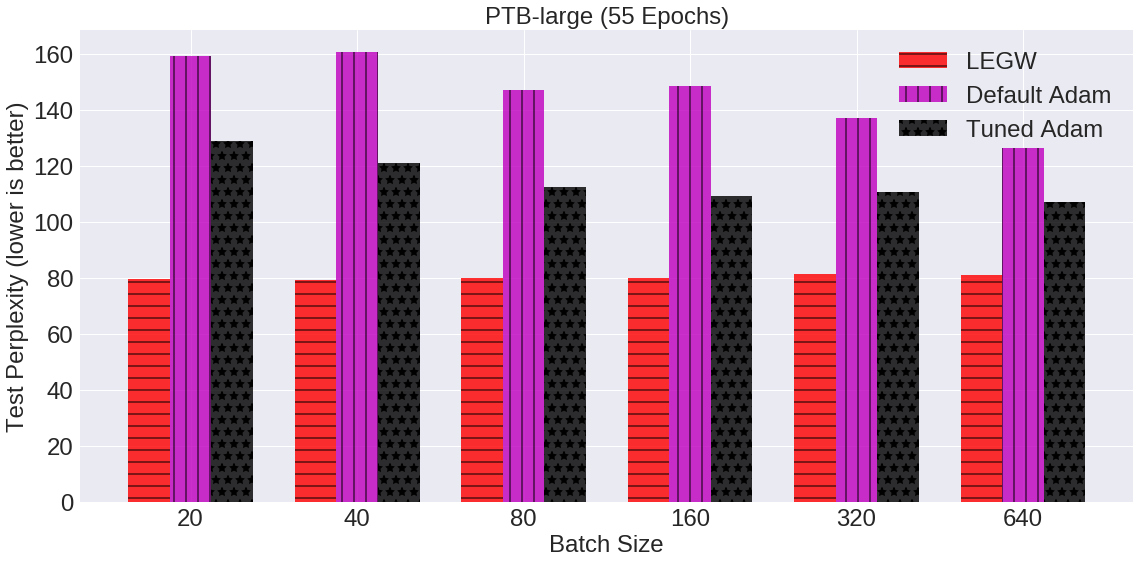}
\label{fig:legw_vs_adam3}}
\subfigure[]{\includegraphics[width=3.4in]{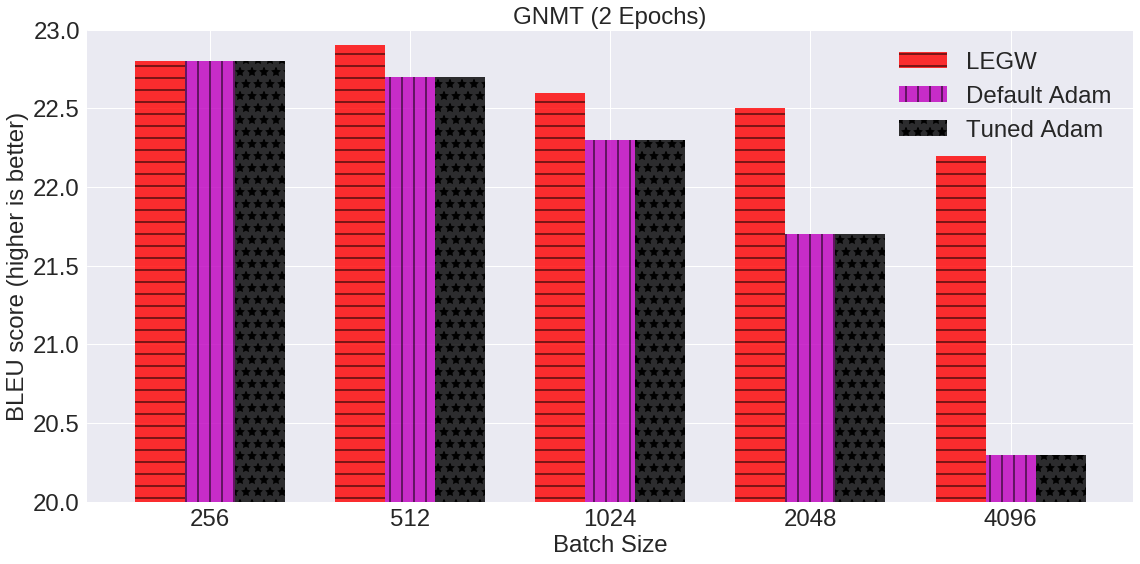}
\label{fig:legw_vs_adam4}}
\vspace{-10pt}\caption{\footnotesize For the perplexity, lower is better. For BLEU score, higher is better. LEGW performs much better than Adam solver for PTB and GNMT applications (Running the same number of epochs). Even we comprehensively tuned the learning rate of Adam, it still can not beat LEGW. \vspace{-10pt}}
\label{fig:legw_vs_adam}
\end{figure}

The comparison between Adam and LEGW is shown in Figure \ref{fig:legw_vs_adam}.
We observe LEGW performs better than Adam for PTB and GNMT applications in the same number of epochs. 
LEGW is also more constant and achieves higher accuracy than Adam for large-batch cases.
It is worth noting that we carefully tuned the learning rate of Adam solver and made sure it gets the best performance. For MNIST application, the tuning space is \{0.0001, 0.0002, 0.0003, ..., 0.0010\}. For PTB application, the tuning space is \{0.001, 0.002, 0.003, ..., 0.020\} and \{0.0001, 0.0002, 0.0003, ..., 0.0020\}. For GNMT application, the tuning space is \{0.001, 0.002, 0.003, ..., 0.020\} and \{0.0001, 0.0002, 0.0003, ..., 0.0020\}. The detailed tuning results can be found from the appendix of this paper.
Therefore, we conclude that LEGW is a better auto-tuning scheme compared to state-of-the-art approaches.

\subsection{Compared to Comprehensive Tuning}
\begin{figure}[ht]
\centering
\renewcommand{\thesubfigure}{\thefigure.\arabic{subfigure}}
\subfigure[]{\includegraphics[width=3.2in]{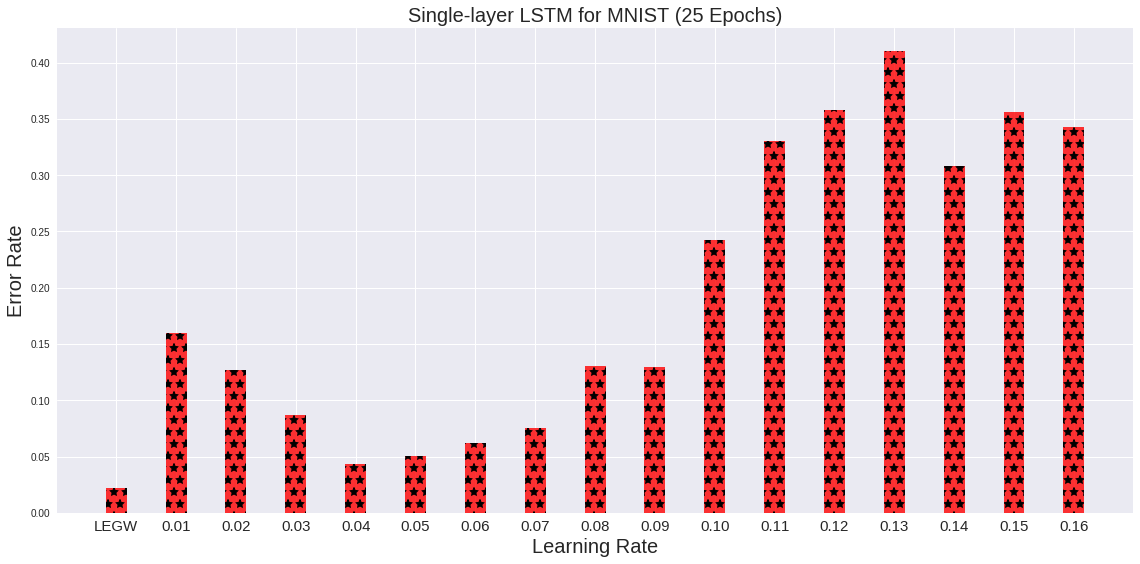}
\label{fig:legw_vs_tuning1}}
\subfigure[]{\includegraphics[width=3.2in]{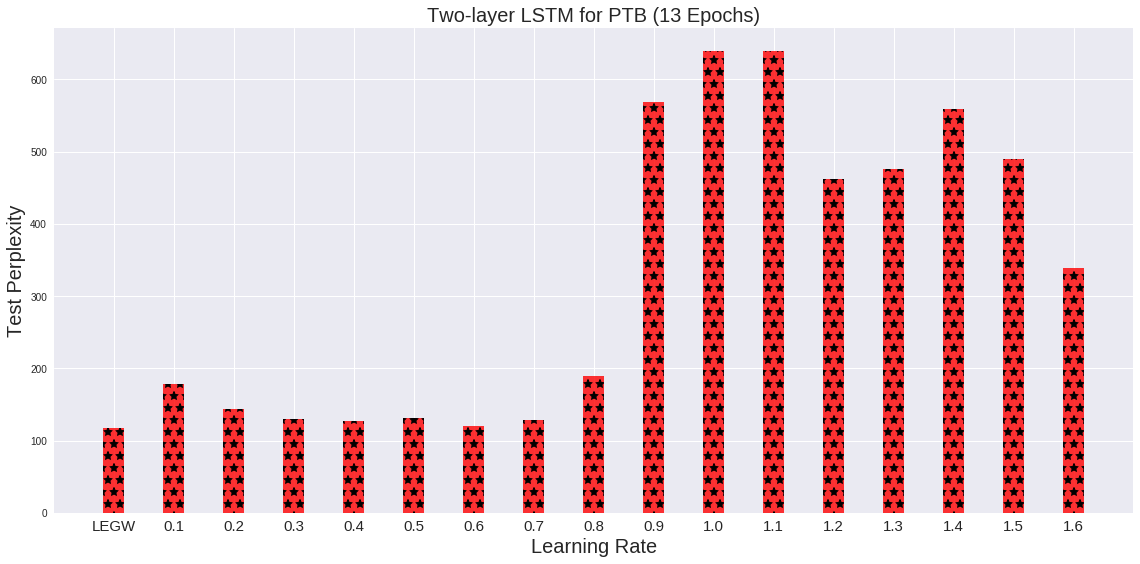}
\label{fig:legw_vs_tuning2}}
\vspace{-10pt}\caption{\footnotesize The data in this figure is collected from 8K batch size. Even when we comprehensively tune the learning rate of the baseline, it still is not able to beat LEGW. For other 
hyper-parameters and learning rate decay schemes, LEGW uses the same setting with the baseline.}
\label{fig:legw_vs_tuning}
\end{figure}

\begin{figure}[ht]
\centering
\vspace{-10pt}\renewcommand{\thesubfigure}{\thefigure.\arabic{subfigure}}
\subfigure[]{\includegraphics[width=3.2in]{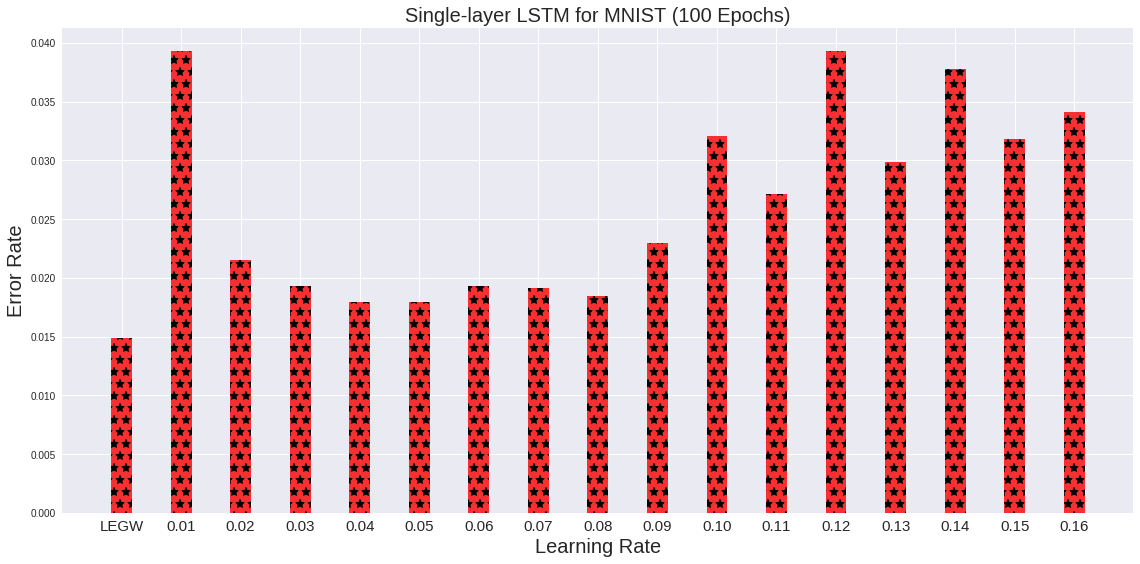}
\label{fig:legw_vs_tuning_long1}}
\subfigure[]{\includegraphics[width=3.2in]{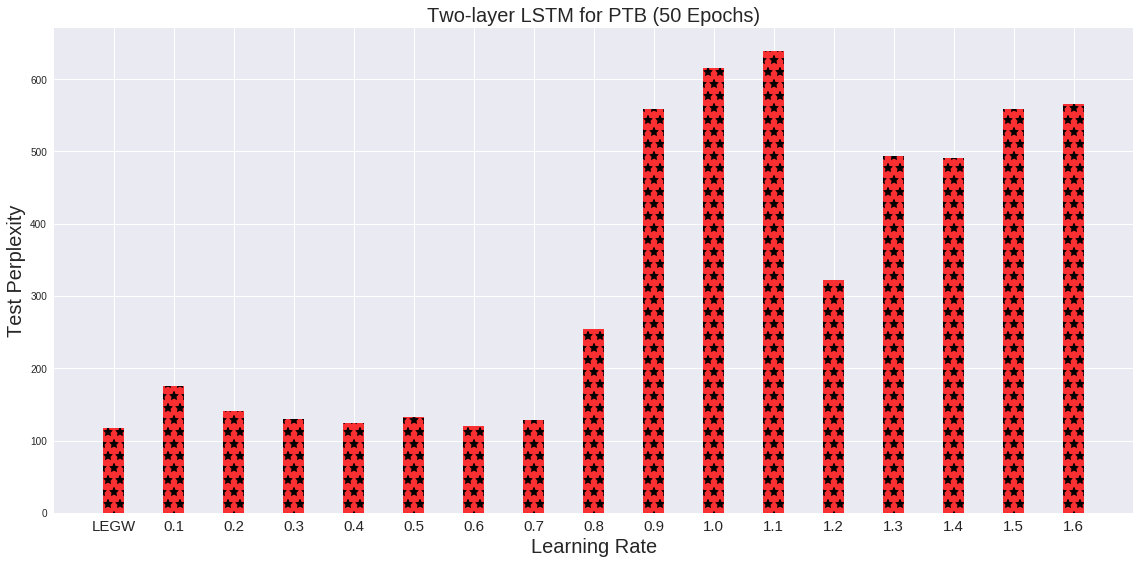}
\label{fig:legw_vs_tuning_long2}}
\vspace{-10pt}\caption{\footnotesize
The data in this figure is collected from 640 batch size. Even we comprehensively tune the initial learning rate of the baseline, it still is not able to beat LEGW. For other hyper parameters and learning rate decay scheme, LEGW uses the same setting with the baseline. Furthermore, we run the training long enough to make sure all of them converge. LEGW is still better.
\vspace{-20pt}}
\label{fig:legw_vs_tuning_long}
\end{figure}

To prove the effectiveness of LEGW, we make a comparison between LEGW and the comprehensive tuning baseline for the largest batch size.
For the MNIST dataset, since the model uses a constant learning rate for momentum solver, we only tune the learning rate.
We comprehensively tune the learning rate and find only the range of [0.01, 0.16] is effective.
After tuning the learning rate from 0.01 to 0.16, we observe that LEGW's accuracy is higher than the best tuned version (Figure \ref{fig:legw_vs_tuning}.1).
For PTB dataset, both the baseline and LEGW use the same exponential learning rate decay scheme.
We comprehensively tune the initial learning rate for baseline and we find only the range from 0.1 to 1.6 is effective.
Then we tune the learning rate within the effective range, the baseline's highest accuracy is still lower than LEGW's accuracy (Figure \ref{fig:legw_vs_tuning}.2).
We also run the training algorithms long enough to make sure all of them converged.
For MNIST dataset, we increase the number of epochs from 25 to 100.
For PTB dataset, we increase the number epochs from 13 to 50.
Even when comprehensive turning versions are allowed to run longer, LEGW is still able to beat them (Figure \ref{fig:legw_vs_tuning_long}).
Furthermore, LEGW does not require hyper-parameter tuning.

\section{ImageNet Training with ResNet-50\label{sec:imagenet}}
To show its robustness, we also apply LEGW to CNN applications.
We use LEGW together with a LARS optimizer \cite{you2017scaling} for ImageNet training with ResNet50.
According to Stanford DAWN benchmark\footnote{https://dawn.cs.stanford.edu/benchmark/}, 93\% top-5 accuracy is the metric of correct ResNet50 implementation.
We are able to scale the batch size to 32K and achieve the target accuracy without tuning hype-parameters (Table \ref{tab:resnet50}). We achieved constant performance and higher accuracy compared to existing tuning schemes (Figure \ref{fig:legw_vs_ls_resnet50}).

\begin{table}
 \scriptsize
  \caption{By using LEGW, we can scale the batch size of GNMT training from 256 to 4K without influencing the BLEU score. The baseline's BLEU score is 21.8. Since linear warmup epochs means fixed the warmup iterations, we set the warmup iterations as 200.
  }
  \label{tab:gnmt}
\centering
    \vspace{3pt}
  \begin{tabular}{|c|c|c|c|c|c|c|}
  \hline
  Batch Size & Init LR & Warmup & Epochs & BLEU\\
    \hline
    \hline
    256 & $2^{-0.5}/10^3$ & 0.0145 epochs & 2 & 22.7\\
   \hline
   512 & $2^{0.0}/10^3$ & 0.0290 epochs & 2 & 22.9\\
   \hline
   1024 & $2^{0.5}/10^3$ & 0.0580 epochs & 2 & 22.6\\
   \hline
   2048 & $2^{1.0}/10^3$ & 0.1160 epochs & 2 & 22.5\\
   \hline
   4096 & $2^{1.5}/10^3$ & 0.2320 epochs & 2 & 22.2\\
   \hline
  \end{tabular}
\end{table}

\begin{table}
 \scriptsize
  \caption{\footnotesize LEGW scales the batch size for ImageNet training by ResNet-50 without tuning hype-parameters. According to Stanford DAWN benchmark, 93\% top-5 accuracy for ImageNet dataset is the metric of correct ResNet50 implementation.
  }
  \label{tab:resnet50}
\centering
    \vspace{3pt}
  \begin{tabular}{|c|c|c|c|c|c|c|}
  \hline
  Batch Size & Init LR & Warmup & Epochs & Top-5 Accuracy\\
    \hline
    \hline
    1024 & $2^{2.5}$ & 10/$2^5$ epochs & 90 & 0.9336\\
   \hline
   2048 & $2^{3.0}$ & 10/$2^4$ epochs & 90 & 0.9325\\
   \hline
   4096 & $2^{3.5}$ & 10/$2^3$ epochs & 90 & 0.9334\\
   \hline
   8192 & $2^{4.0}$ & 10/$2^2$ epochs & 90 & 0.9355\\
   \hline
   16384 & $2^{4.5}$ & 10/$2^1$ epochs & 90 & 0.9343\\
   \hline
    32768 & $2^{5.0}$ & 10 epochs & 90 & 0.9318\\
   \hline
  \end{tabular}
\end{table}

\section{Speedup\label{sec:speedup}}
For ImageNet training with ResNet50, our auto-tuning approach is able to scale the batch size to 32K without losing accuracy.
On a TPU-v2 Pod, we are able to finish the training in 7 minutes.
The baseline \cite{goyal2017accurate} can only scale the batch size to 8K, which takes 16 minutes on the same TPU-v2 Pod.
Ying et al. \cite{ying2018image} are able to finish the ImageNet training with ResNet-50 in 2.2 minutes by LARS solver \cite{you2017scaling}; however,
they use a better hardware (i.e. TPU-v3 Pod).
If we port their code to TPU-v2 Pod, they achieve the same speed as us.
The difference between our results and theirs is that they need to tune the hyper-parameters manually. We use an auto-tuning technique (i.e. LEGW).

For the other four LSTM-based applications, LEGW can also help the system utilize a much larger batch size without sacrificing accuracy. This leads to significant speedups on all the four datasets (Figure \ref{fig:legw_speed}).
For example, our GNMT baseline with a batch size of 256 needs more than 2 hours to finish the training on a cloud TPU-v2.
With LEGW, the GNMT with a batch size of 4096 can finish the training in 33 minutes on the same cloud TPU-v2.
In summary, LEGW achieves a 5.3$\times$ average speedup over the baselines for 4 LSTM-based applications on the same hardware.

\section{Conclusion}
LEGW is an auto-tuning method based on a warmup technique and Sqrt Scaling scheme.
For LSTM applications, we are able to scale the batch size by a factor of 64$\times$ without losing accuracy and without tuning the hyper-parameters.
For CNN applications, LEGW is able to keep accuracy constant even as we scale the batch size to 32K, and  
we have demonstrated that LEGW works uniformly better than previous large-batch auto-tuning techniques (Figure \ref{fig:legw_vs_ls_resnet50}).
For four LSTM applications, while running on the same hardware, LEGW achieves a 5.3$\times$ average speedup.
We also provide a theoretical explanation for the effectiveness of LEGW.

\section{Acknowledge}
We would like to thank the comments from George E. Dahl at Google.
We would like to thank Sameer Kumar and Tao Wang at Google for helping us implement LARS algorithm on TPU Pod.
This technical report is only a research paper.
The TPU's speed in this paper should not be considered as Google's official number.
The readers should check Google's official documents for TPU's speed. 
The content, views and conclusions presented in this paper do not necessarily reflect the position of Google.



\clearpage

{\footnotesize
\bibliographystyle{abbrv}
\bibliography{mybib}  
}

\section{Appendix}
\begin{figure}[ht]
\centering
\renewcommand{\thesubfigure}{\thefigure.\arabic{subfigure}}
\subfigure[]{\includegraphics[width=3.3in]{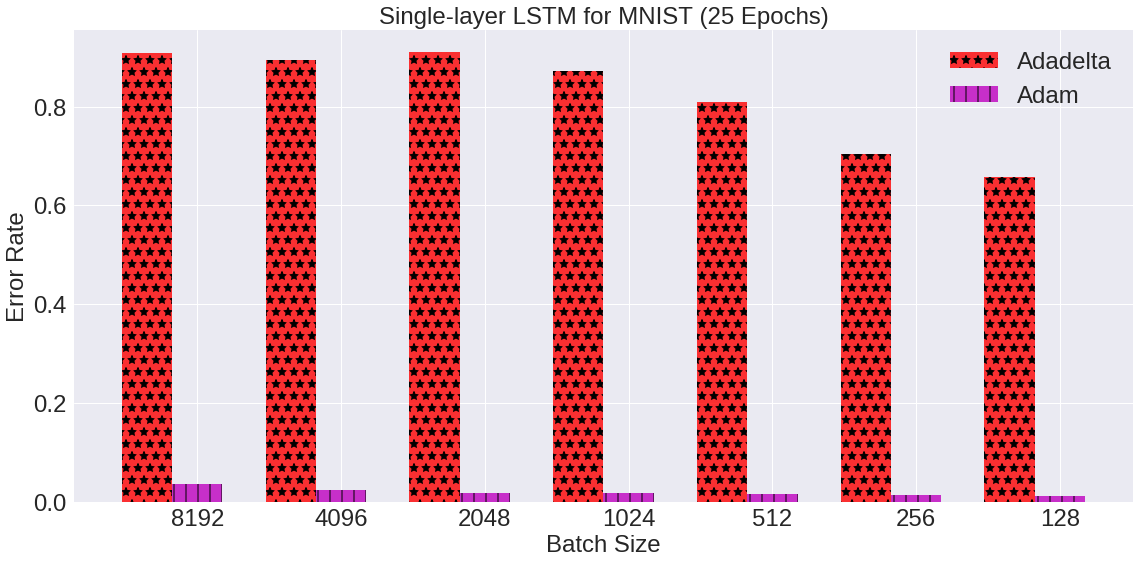}
\label{fig:adam_adadelta1}}
\subfigure[]{\includegraphics[width=3.3in]{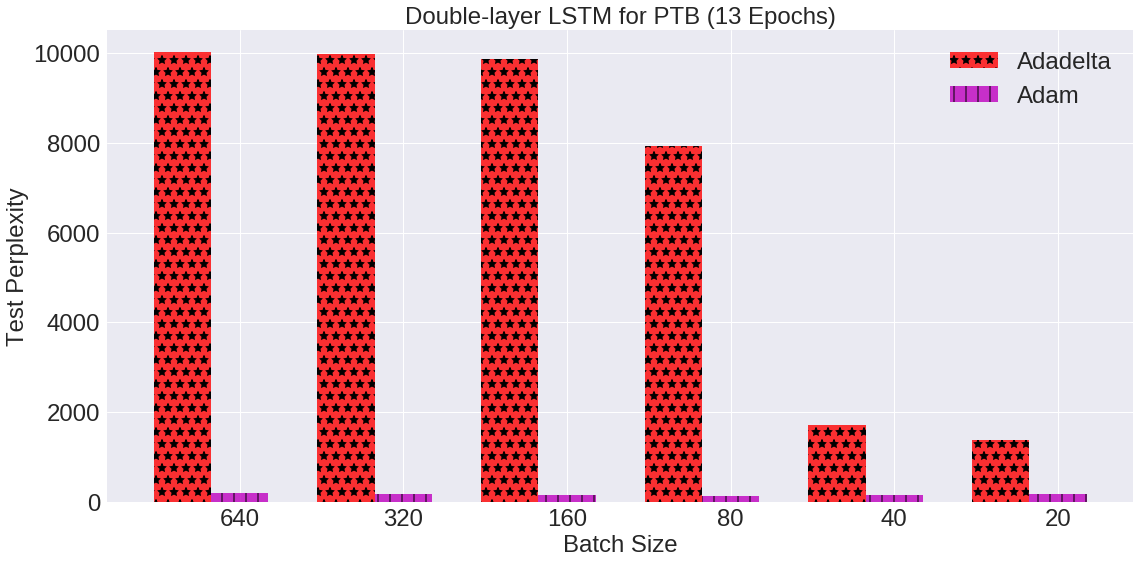}
\label{fig:adam_adadelta2}}
\caption{By just using the default hyper parameters, we find Adam works much better than Adadelta for MNIST and PTB datasets.}
\label{fig:adam_adadelta}
\end{figure}

\begin{figure}[ht]
\vspace{-10pt}\centering
\renewcommand{\thesubfigure}{\thefigure.\arabic{subfigure}}
\subfigure[]{\includegraphics[width=3.3in]{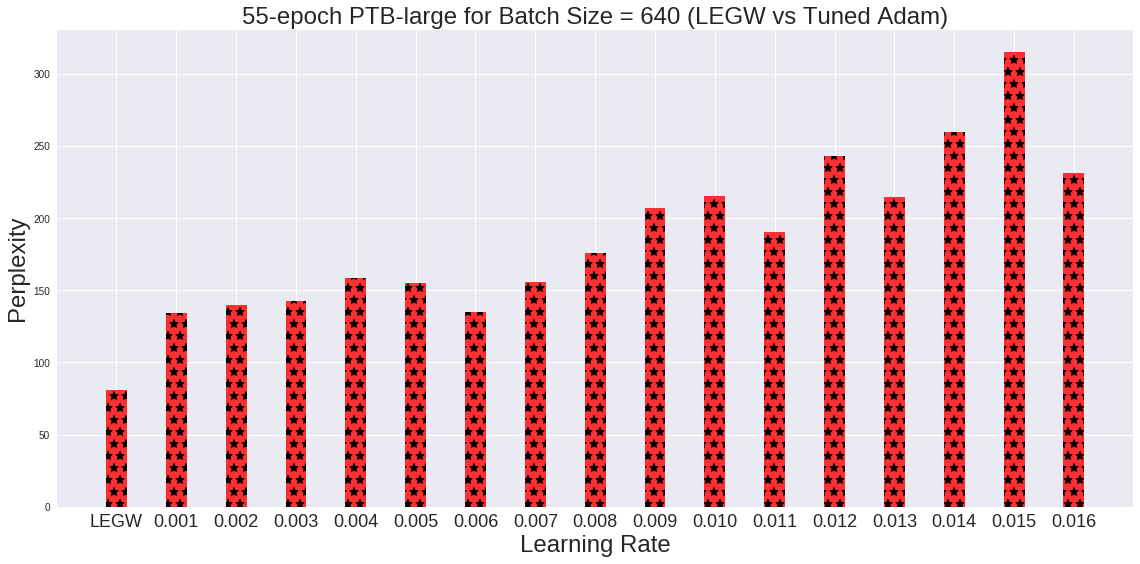}
\label{fig:legw_visualize_1}}
\subfigure[]{\includegraphics[width=3.3in]{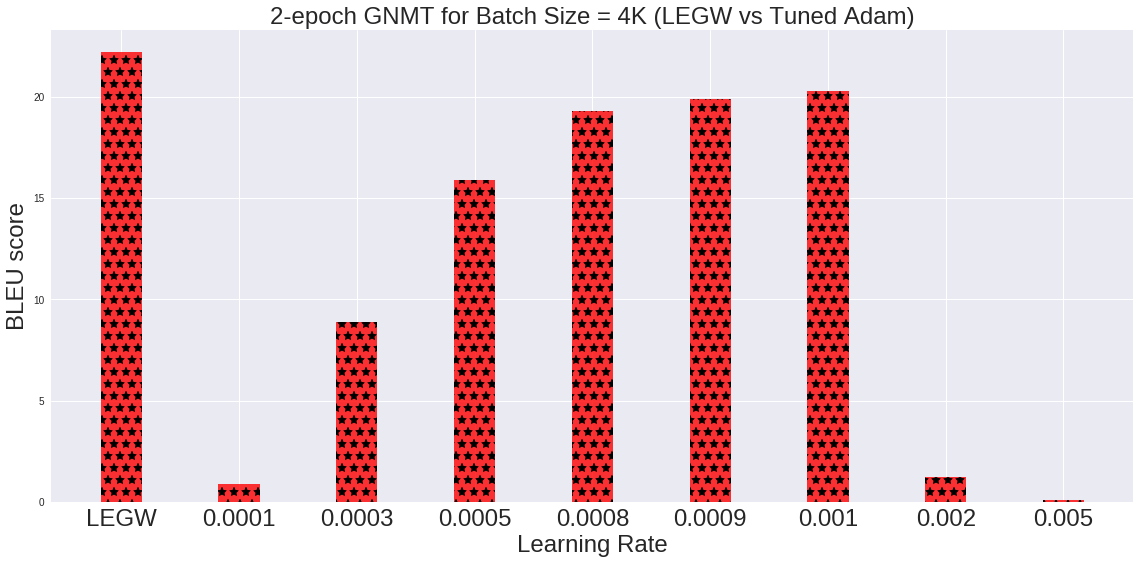}
\label{fig:legw_visualize_2}}
\vspace{-5pt}\caption{\footnotesize For Perplexity, lower is better. For BLEU score, higher is better. From these figures, we can observe that LEGW performs better than the tuned Adam solver for PTB-large and GNMT applications.\vspace{-10pt}}
\label{fig:legw_vs_adam_app}
\end{figure}

For MNIST and PTB datasets, we observe Adam performs much better than Adadelta (Figure \ref{fig:adam_adadelta}).
Figure \ref{fig:legw_vs_adam_app} shows that LEGW performs better than the tuned Adam solver for PTB-large and GNMT applications.

\end{document}